\documentclass[runningheads]{llncs}
\usepackage{times}
\usepackage{helvet}
\usepackage{amsmath} 
\usepackage{graphicx}
\DeclareGraphicsExtensions{.pdf,.png,.jpg,.eps}
\usepackage{color}
\usepackage{listings}
\usepackage{wrapfig}
\def\ar{\leftarrow}
\def\beq{\begin{equation}}
\def\eeq#1{\label{#1}\end{equation}}
\def\ba{\begin{array}}
\def\ea{\end{array}}
\def\bi{\begin{itemize}}
\def\ei{\end{itemize}}
\def\i#1{\hbox{\it #1\/}}

\def\no{\i{not}}

\def\ar{\leftarrow}
\def\rar{\rightarrow}
\def\lrar{\leftrightarrow}

\def\sm{\hbox{\rm SM}}

\def\dg{\hbox{\rm DG}}

\def\fsm{\hbox{\rm SM}}

\def\mi#1{\mathit{#1}}
\def\proof{\noindent{\bf Proof}.\hspace{3mm}}
\def\qed{\quad \vrule height7.5pt width4.17pt depth0pt \medskip}

\def\bC{{\bf{c}}}
\def\vbC{{\wh{\bf{c}}}}

\def\wh{\widehat}
\def\mvis{\!=\!}

\def\true{\hbox{\sc true}}

\long\def\nb#1{}

\long\def\comment#1{}
\long\def\COMMENT#1\ENDCOMMENT{\message{(Commented text...)}\par}
\long\def\CMT#1\EOC{\message{(Commented text...)}}
\long\def\BOC#1\EOC{\message{(Commented text )}}
\long\def\BOCC#1\EOCC{\message{(Commented text )}}
\long\def\BOCCC#1\EOCCC{\message{(Commented text )}}
\long\def\optional#1{}

\newtheorem{prop}{Proposition}
\newtheorem{thm}{Theorem}

\newcommand{\cblu}{\color{blue}}
\newcommand{\cbla}{\color{black}}

\title{System {\sc aspmt2smt}: \\ Computing ASPMT Theories by SMT Solvers}

\author{Michael Bartholomew \and Joohyung Lee}
\institute{
School of Computing, Informatics, and Decision Systems Engineering \\
Arizona State University, Tempe, USA \\
{\tt \{mjbartho,joolee\}@asu.edu}}

\begin{document}

\maketitle

\begin{abstract}
Answer Set Programming Modulo Theories (ASPMT) is an approach to
combining answer set programming and satisfiability modulo theories
based on the functional stable model semantics. It is shown that the
tight fragment of ASPMT programs can be turned into SMT instances,
thereby allowing SMT solvers to compute stable models of ASPMT
programs. In this paper we present a compiler called {\sc aspsmt2smt},
which implements this translation. The system uses ASP grounder {\sc
 gringo} and SMT solver {\sc z3}. {\sc gringo} partially grounds
input programs while leaving some variables to be processed by {\sc
  z3}. We demonstrate that the system can effectively handle real
number computations for reasoning about continuous changes. 
\end{abstract} 

\section{Introduction}

Answer Set Programming (ASP) is a widely used declarative computing
paradigm. Its success is largely due to the expressivity of its
modeling language and efficiency of ASP solvers thanks to intelligent
grounding and efficient search methods that originated from
propositional satisfiability (SAT) solvers.
While grounding methods implemented in ASP solvers are highly
optimized, ASP inherently suffers when variables range over large
domains. Furthermore, real number computations are not supported by
ASP solvers because grounding cannot be even applied. Thus reasoning
about continuous changes even for a small interval requires loss of
precision by discretizing the domain. 

Satisfiability Modulo Theories (SMT) emerged as an
enhancement of SAT, which can be also viewed as a special case of
(decidable) first-order logic in which certain predicate and function
symbols in background theories have fixed interpretations. Example
background theories are the theory of real numbers, the theory of
linear arithmetic, and difference logic.

A few approaches to loosely combining ASP and SMT/CSP exist
\cite{gebser09constraint,balduccini09representing,janhunen11tight},
in which nonmonotonicity of the semantics is related to predicates in
ASP but has nothing to do with functions in SMT/CSP. For instance, while 
\[
  \i{WaterLevel}(t\!+\!1,\i{tank}, l)\ \ar\ \i{WaterLevel}(t, \i{tank}, l),\ 
     \no\ \neg \i{WaterLevel}(t\!+\!1, \i{tank}, l)
\]
($t$ is a variable ranging over steps; $l$ is a variable for the water level) 
represents the default value of water level correctly (albeit grounding
suffers when the variables range over a large numeric domain), 
rewriting it in the language of {\sc clingcon}---a combination of ASP
solver {\sc clingo} and constraint solver {\sc gecode}---as 
\[
  \i{WaterLevel}(t\!+\!1, \i{tank})\!=^\$\! l\ \ar\
       \i{WaterLevel}(t,\i{tank})\!=^\$\! l,\ 
             \no\ \neg(\i{WaterLevel}(t\!+\!1,\i{tank})\!=^\$\! l) 
\]
does not express the concept of defaults correctly. 
\BOCC
A more detailed comparison between our approach and these loose integrations of ASP with SMT or CSP can be found in~\cite{bartholomew13functional} but
the inability for those integrations to express the notion of defaults
as above is due to the semantic understanding of functions as in 
classical logic rather than under a nonmonotonic semantics as is the
case with ASPMT. Instead, these formalisms must represent this notion
the same way as in classical logic--either using additional auxiliary 
abnormality constants or in a way that preserves the signature but is 
not elaboration tolerant.
\EOCC

In~\cite{bartholomew13functional}, it was observed that a tight
integration of ASP and SMT requires a generalization of the
stable model semantics in which default reasoning can be expressed
via (non-Herbrand) functions as well as predicates. Based on the
functional stable model semantics from~\cite{bartholomew12stable}, a
new framework called ``Answer Set Programming Modulo Theories (ASPMT)'' was proposed, which is analogous to SMT. 
Just like SMT is a generalization of SAT and, at the same time, a
special case of first-order logic with fixed background theories, 
ASPMT is a generalization of the traditional ASP and,
at the same time, a special case of the functional stable model
semantics in which certain background theories are assumed. Unlike
SMT, ASPMT allows expressive nonmonotonic reasoning as allowed in ASP.

It is shown in~\cite{bartholomew13functional}, a fragment of
ASPMT instances can be turned into SMT instances, so that SMT solvers
can be used for computing stable models of ASPMT instances. 
In this paper, we report an implementation of this translation in the
system called ``{\sc aspmt2smt}.''  
The system first partially grounds the theory by replacing ``ASP variables'' with ground terms, leaving other ``SMT variables'' ungrounded. Then, it computes the completion
of the theory. Under certain conditions guaranteed by the class of
ASPMT theories considered, the remaining variables can then be
eliminated. After performing this elimination, the {\sc aspmt2smt} system
then invokes the {\sc z3} system to compute classical models, which
correspond to the stable models of the original theory.
We show that several examples involving both discrete changes as
well as continuous changes can be naturally represented in the input
language of {\sc aspmt2smt}, and can be effectively computed. 
\cbla

The paper is organized as follows. In section~\ref{sec:prelim}, we
first review the functional stable models semantics and as its special
case, ASPMT, and then review the theorem on completion
from~\cite{bartholomew12stable}. 
In section~\ref{sec:variable}, we describe the
process of variable elimination used by the system. In
section~\ref{sec:system}, we describe the architecture of the system as
well as the syntax of the input language. Finally, in section~\ref{sec:experiment}, we present several
experiments with and without continuous reasoning and compare the
performance to ASP solver {\sc clingo} when appropriate.

The system is available at {\tt \url{http://reasoning.eas.asu.edu/aspmt}}.

\section{Preliminaries}\label{sec:prelim}

\subsection{Review of the Functional Stable Model Semantics}

We review the stable model semantics of intensional functions
from~\cite{bartholomew12stable}.
Formulas are built the same as in first-order logic.

Similar to circumscription, for predicate symbols (constants or
variables) $u$ and $c$, expression $u\le c$ is defined as shorthand for
\hbox{$\forall {\bf x}(u({\bf x})\rar c({\bf x}))$}.
Expression $u=c$ is defined as
\hbox{$\forall {\bf x}(u({\bf x})\lrar c({\bf x}))$}
if $u$ and $c$ are predicate symbols, and
$\forall {\bf x}(u({\bf x})=c({\bf x}))$
if they are function symbols.
For lists of symbols ${\bf u}=(u_1,\dots,u_n)$ and
${\bf c}=(c_1,\dots,c_n)$, expression ${\bf u}\le {\bf c}$ is defined
as $(u_1\le c_1)\land\dots\land (u_n\le c_n)$, 
and similarly, expression ${\bf u} = {\bf c}$ is defined
as $(u_1 = c_1)\land\dots\land (u_n = c_n)$.
Let $\bC$ be a list of distinct predicate and function constants, and
let $\vbC$ be a list of distinct predicate and function variables
corresponding to~$\bC$. 
By $\bC^\mi{pred}$ ($\bC^\mi{func}$, respectively) we mean the
list of all predicate constants (function constants, respectively) in
$\bC$, and by $\vbC^\mi{pred}$ ($\vbC^\mi{func}$, respectively) the
list of the corresponding predicate variables (function
  variables, respectively) in $\vbC$.

For any formula $F$ and any list of predicate and function constants $\bC$, which we call {\sl intensional} constants, expression $\fsm[F;\ \bC]$ is defined as
\[
   F\land\neg\exists \vbC(\vbC<\bC\land F^*(\vbC)),
\]
where $\vbC<\bC$ is shorthand for 
$(\vbC^\mi{pred}\le \bC^\mi{pred})\land\neg (\vbC=\bC)$, 
and $F^*(\vbC)$ is defined recursively as follows.
\begin{itemize}
\item When $F$ is an atomic formula, $F^*$ is $F'\land F$ where $F'$
  is obtained from $F$ by replacing all intensional (function and
  predicate) constants ${\bf c}$ in it with the corresponding
  (function and predicate) variables from ${\vbC}$;


\item  $(G\land H)^* = G^*\land H^*$;\ \ \ \ \
       $(G\lor H)^* = G^*\lor H^*$;

\item  $(G\rar H)^* = (G^*\rar H^*)\land (G\rar H)$;

\item  $(\forall x G)^* = \forall x G^*$;\ \ \ \ \
       $(\exists xF)^* = \exists x F^*$.
\end{itemize}
(We understand $\neg F$ as shorthand for $F\rar\bot$; $\top$ as
$\neg \bot$; and $F\lrar G$ as $(F\rar G)\land (G\rar F)$.)

When $F$ is a sentence, the models of $\fsm[F; {\bf c}]$ are called
the {\em stable} models of~$F$ {\em  relative to} ${\bf c}$. They are
the models of~$F$ that are ``stable'' on ${\bf c}$.
The definition can be easily extended to formulas of many-sorted
signatures.

This definition of a stable model is a proper generalization of the
one from~\cite{ferraris11stable}, which views logic programs as a
special case of first-order formulas. 

We will often write $G\ar F$, in a rule form as in logic programs, 
to denote the universal closure of $F\rar G$. 
A finite set of formulas is identified with the conjunction of the
formulas in the set.

By $\{c\mvis v\}$, we abbreviate the formula $c\mvis v\lor \neg (c\mvis v)$
which, in the functional stable model semantics, can be intuitively
understood as ``by default, $c$ is mapped to $v$''. 

\subsection{ASPMT as a Special Case of the Functional Stable Model
  Semantics}  \label{ssec:aspmt}
We review the semantics of ASPMT described in~\cite{bartholomew13functional}.
Formally, an SMT instance is a formula in many-sorted first-order
logic, where some designated function and predicate constants are
constrained by some fixed background interpretation. SMT is the
problem of determining whether such a formula has a model that expands
the background interpretation~\cite{barrett09satisfiability}.

The syntax of ASPMT is the same as that of SMT. Let $\sigma^{bg}$ be
the (many-sorted) signature of the background theory~$bg$. An
interpretation of $\sigma^{bg}$ is called a {\em background
  interpretation} if it satisfies the background theory. For instance,
in the theory of reals, we assume that $\sigma^{bg}$ contains the set
$\mathcal{R}$ of symbols for all real numbers, the set of arithmetic
functions over real numbers, and the set $\{<, >, \le, \ge, =\}$ of
binary predicates over real numbers. Background interpretations
interpret these symbols in the standard way.

Let $\sigma$ be a signature that is disjoint from $\sigma^{bg}$.
We refer to functions in $\sigma^{bg}$ as interpreted functions and
functions in $\sigma$ as uninterpreted functions.
We say that an interpretation $I$ of $\sigma$ satisfies $F$
w.r.t. the background theory $bg$, denoted by $I\models_{bg} F$,
if there is a background interpretation $J$ of $\sigma^{bg}$ that has
the same universe as $I$, and $I\cup J$ satisfies $F$.
For any ASPMT sentence $F$ with background theory
$\sigma^{bg}$, interpretation $I$ is a {\em stable model} of $F$ relative
to~${\bf c}$ (w.r.t. background theory $\sigma^{bg}$) if
$I\models_{bg} \sm[F; \bC]$. When $\bC$ is empty, the stable models of
$F$ coincides with the models of $F$.


Consider the following running example from a Texas Action Group
discussion\footnote{http://www.cs.utexas.edu/users/vl/tag/continuous\_problem}.

\begin{quote}
A car is on a road of length $L$.  If the accelerator is activated, the
car will speed up with constant acceleration $A$ until the accelerator is
released or the car reaches its maximum speed $MS$, whichever comes first.
If the brake is activated, the car will slow down with acceleration
$\neg A$
until the brake is released or the car stops, whichever comes first.
Otherwise, the speed of the car remains constant.
Give a formal representation of this domain, and write a program that
uses your representation to generate a plan satisfying the following
conditions:  at duration 0, the car is at rest at one end of the road; at
duration $T$, it should be at rest at the other end.
\end{quote}

This problem is an instance of planning with continuous time, which
requires real number computations. 

The domain can be naturally represented in ASPMT as follows.
Below $s$ ranges over time steps, $b$ is a boolean variable,
$x,y,a,c,d$ are all real variables, and ${\rm A}$ and ${\rm MS}$
are some specific numbers.

We represent that the actions $\i{Accel}$ and $\i{Decel}$ are
exogenous and the duration of each time step is to be arbitrarily
selected as
\[
\ba{l}
\{\i{Accel}(s) = b\}, \hspace{1cm} 
\{\i{Decel}(s) = b\}, \hspace{1cm}
\{\i{Duration}(s) = x\}.\\
\ea
\]
Both $\i{Accel}$ and $\i{Decel}$ cannot be performed at the
same time: 
\[
\ba{l}

\bot \leftarrow \i{Accel}(s) = \true \land \i{Decel}(s) = \true.\\
\ea
\]
The effects of $\i{Accel}$ and $\i{Decel}$ on $\i{Speed}$ are described as
\[
\ba{rl}
  \i{Speed}(s+1) = y \ar& \i{Accel}(s)\mvis\true\ \land\ 
                          \i{Speed}(s)\mvis x\ \land\ 
                          \i{Duration}(s)\mvis d  \\
   & \land\ (y = x+{\rm A}\times d),  \\
  \i{Speed}(s+1) = y \ar& \i{Decel}(s)\mvis\true\ \land\ 
                          \i{Speed}(s)\mvis x\ \land\ 
                          \i{Duration}(s)\mvis d\  \\
   & \land\ (y = x-{\rm A}\times d). 
\ea
\]
The preconditions of $\i{Accel}$ and $\i{Decel}$ are described as
\[
\ba{rl}
\bot \ar & \i{Accel}(s)\mvis\true\ \land\ 
           \i{Speed}(s)\mvis x\ \land\
           \i{Duration}(s)\mvis d \\
         & \land\ (y = x+{\rm A}\times d)\ \land\ (y > {\rm MS}), \\
\bot \ar & \i{Decel}(s)\mvis\true\ \land\ 
           \i{Speed}(s)\mvis x\ \land\
           \i{Duration}(s)\mvis d \\
         & \land\ (y = x-{\rm A}\times d)\ \land\ (y < 0).
\ea
\]
$\i{Speed}$ is inertial:
\[
  \{\i{Speed}(s+1) = x\} \ar \i{Speed}(s) = x.
\]
The $\i{Location}$ is defined in terms of $\i{Speed}$ and
$\i{Duration}$ as 
\[
\ba{rl}
\i{Location}(s+1) = y \ar & \i{Location}(s) = x \land 
     \i{Speed}(s)=a  \land 
     \i{Speed}(s+1) = c\  \\
     & \land\ \i{Duration}(s)\mvis d\ \land\ 
     y = x + ((a + c) / 2)\times d. 
\ea
\]

\subsection{Theorem on Completion}\label{ssec:comp}

We review the theorem on completion from~\cite{bartholomew13functional}.
The {\em completion} turns  ``tight'' ASPMT instances into 
equivalent SMT instances, so that SMT solvers can be used for
computing this fragment of ASPMT. 

We say that a formula $F$ is in {\em Clark normal form} (relative to
the list ${\bf c}$ of intensional constants) if it is a conjunction of
sentences of the form 
\beq
   \forall {\bf x} (G\rar p({\bf x}))
\eeq{cnf-p}
and 
\beq
\forall {\bf x}y (G \rar f({\bf x}) \mvis y)
\eeq{cnf-f}
one for each intensional predicate $p$ and each intensional function
$f$, where ${\bf x}$ is a list of distinct object variables, $y$ is
a variable, and $G$ is a formula that has no free variables other than
those in ${\bf x}$ and $y$, and sentences of the form
\beq
  \ar G.
\eeq{cnf-constr}

The {\em completion} of a formula $F$ in Clark normal form (relative
to ${\bf c}$) is obtained from~$F$ by replacing each conjunctive term
(\ref{cnf-p}) with
\beq
  \forall {\bf x} (p({\bf x})\lrar G),
\eeq{comp-p}
each conjunctive term (\ref{cnf-f}) with 
\beq
  \forall {\bf x}y (f({\bf x})\mvis y\lrar G),
\eeq{comp-f}
and each conjunctive term \eqref{cnf-constr} with $\neg G$.

An occurrence of a symbol or a subformula in a formula $F$ is called
{\em strictly positive} in~$F$ if that occurrence is not in the
antecedent of any implication in $F$ .

The {\em dependency graph} of a formula $F$ relative to ${\bf c}$,
denoted by $\dg_\bC[F]$, is the directed graph that 
\begin{itemize}
\item  has all members of ${\bf c}$ as its vertices, and
\item  has an edge from $c$ to $d$ if, for some strictly positive
  occurrence of $G\rar H$ in $F$, $c$ has a strictly positive
  occurrence in~$H$, and $d$ has a strictly positive occurrence in~$G$. 
\end{itemize}
We say that $F$ is {\em tight} on {\bf c} if the dependency graph of
$F$ relative to {\bf c} is acyclic. 
\BOCC
\cblu
For example, 
\[
   ((p\rar q)\rar r)\rar p
\]
is tight on $\{p,q,r\}$ because its dependency graph has only one
edge, which goes from $p$ to $r$. On the other hand, the formula is
not tight according to~\cite{ferraris11stable} because, according to
the definition of a dependency graph in that paper, there is an 
additional edge that goes from $p$ to itself. 
\cbla
\EOCC


\begin{thm} (\cite[Theorem~2]{bartholomew13functional})\label{thm:completion}\optional{thm:completion}
For any formula $F$ in Clark normal form that is tight on ${\bf c}$,
an interpretation $I$ that satisfies $\exists xy(x \ne y)$ is a model
of $\sm[F;{\bf c}]$ iff $I$ is a model of the completion of $F$
relative to ${\bf c}$.
\end{thm}

For example, the car example formalization contains the following
implications for the function $\i{Speed}(1)$: 
{\small
\[
\ba {rl}
  \i{Speed}(1)\mvis y\ \ar\ & \i{Accel}(0)\mvis\true\ \land\ 
                          \i{Speed}(0)\mvis x\ \land\ 
                          \i{Duration}(0)\mvis d\  
    \land (y = x+{\rm A}\times d)\ \\ 
  \i{Speed}(1)\mvis y\ \ar\ & \i{Decel}(0)\mvis\true\ \land\ 
                          \i{Speed}(0)\mvis x\ \land\ 
                          \i{Duration}(0)\mvis d\  
   \land (y = x-{\rm A}\times d)\ \\ 
   \i{Speed}(1)\mvis y\ \ar\ & \i{Speed}(0)\mvis y\ \land\
   \neg\neg(\i{Speed}(1)\mvis y)
\ea 
\]
}
($\{c\mvis v\}\ar G$ is strongly equivalent to $c\mvis v\ar G\land
\neg\neg (c\mvis v)$)
and the completion contains the following equivalence.
{\small
\beq
\ba{l}
\i{Speed}(1)  = y\  \lrar\ \\
~~\exists xd(~~ 
      (\i{Accel}(0)\mvis\true\ \land\ 
      \i{Speed}(0)\mvis x\ \land\ 
      \i{Duration}(0)\mvis d\  
      \land\ (y = x+{\rm A}\times d))\ \\ 
~~~~~~~\lor (\i{Decel}(0)\mvis\true\ \land\ 
      \i{Speed}(0)\mvis x\ \land\ 
      \i{Duration}(0)\mvis d\  
     \land\ (y = x-{\rm A}\times d))\   \\
~~~~~~~\lor \i{Speed}(0)=y ~~)
\ea
\eeq{ex:comp}
}

\BOCC
\cblu
If we drop the requirement that $I$ satisfy $\exists xy(x \ne y)$, the statement is no longer correct; consider the interpretation $I$ with universe $\{1\}$ which maps the only function $f$ in the signature to $1$. Take $F$ to be $\top$. $I$ is a model of $\sm[F;{\bf c}]$ but is not a model of the completion, which is equivalent to $\forall y (f = y \leftrightarrow \bot)$.
\cbla
\EOCC

\section{Variable Elimination}\label{sec:variable}

Some SMT solvers do not support variables at all (e.g. iSAT) while
others suffer in performance when handling  variables (e.g. {\sc z3}). 
While we can partially ground the input theories, some variables have large
(or infinite) domains and should not (or cannot) be grounded. Thus, we
consider two types of variables: {\em ASP variables}---variables
which are grounded by ASP grounders---and {\em SMT variables}---variables
which should not be grounded. After eliminating ASP variables by
grounding, we consider the problem of equivalently rewriting the
completion of the partially ground ASPMT theory 
so that the result contains no variables. 

\BOCC
In general, variable elimination is
not a simple task and may be impossible. 
For example, 

$$f = x/3 \leftarrow g = x*x$$
cannot have $x$ eliminated. This is due to the arithmetic expressions
not being invertible( e.g. $f = 3$ means that $x$ can be one of 9, 10,
or 11 and $g = 9$ means that $x$ can be $3$ or $-3$). However,
inverting arithmetic expressions is not always necessary to remove
variables; if we instead have 
$$f = x/3 \leftarrow g = x*x \land h = x$$
then we can substitute $h$ for $x$ to equivalently write
$$f = h/3 \leftarrow g = h*h \land h = h.$$
\EOCC

To ensure that variable elimination can be performed, we impose some
syntactic restrictions on ASPMT instances. We first impose that no 
SMT variable appears in the argument of an uninterpreted function.

We assume {\sc aspmt2smt} programs comprised of rules of the form 
$H \ar B$ where 
\bi
\item  $H$ is $\bot$ or an atom of the form $f({\bf t})=v$, where
  $f({\bf t})$ is a term and $v$ is a variable;
\item  $B$ is a conjunction of atomic formulas possibly preceded with
  $\neg$.
\ei


We define the {\em variable dependency graph} of a conjunction of
possibly negated atomic formulas $C_1 \land \dots \land C_n$ as
follows. The vertices are the variables occurring in $C_1
\land \dots \land C_n$. There is a directed edge from $v$ to $u$ if
there is a $C_i$ that is $v = t$ or $t = v$ for some term $t$ such that $u$
appears in $t$. 
We say a variable $v$ {\em depends on} a variable $u$ if
there is a directed path from $v$ to $u$ in the variable dependency
graph. 
We say a rule $H \ar B$ is  {\em variable isolated} if every
variable $v$ in it occurs in an equality $t = v$ or $v = t$ that is
not negated in $B$ and the variable dependency graph of $B$ is acyclic.

\begin{example}
The rule $f = x \ar g = 2\times x$ is not variable isolated because variable
$x$ does not occur in an equality $x = t$ or $t = x$ in the
body. Instead, we write this as $f = x \leftarrow (g = y)\land (y =
2\times x)$, which is variable isolated.

The rule $f = x \ar (2\times x = y)\land (2\times y = x)$ is not variable isolated;
although variable $y$ occurs in an equality of the form $t = y$, the
dependency graph is not acyclic. 
\end{example}
\cbla

The variable elimination is performed modularly so the process needs 
only to be described for a single equivalence. If an ASPMT program
contains no variables in arguments of uninterpreted functions, any
equivalence in the completion of the ASPMT program will be
of the form 
$$\forall v(f = v \leftrightarrow \exists {\bf x} (B_1(v,{\bf x}) \lor \dots \lor B_k(v,{\bf x})))$$
where each $B_i$ is a conjunction of possibly negated literals and has
$v = t$ as a non-negated subformula, and the variable
dependency graph of $B$ is acyclic. 
In the following, the notation $F^v_t$ denotes the formula
obtained from $F$ by replacing every occurrence of the variable $v$
with the term $t$. We define the process of eliminating variables from
such an equivalence $E$ as follows.

\begin{enumerate}
\item Given an equivalence $E = \forall v(f = v \leftrightarrow \exists {\bf x} (B_1(v,{\bf x}) \lor \dots \lor B_k(v,{\bf x})))$,
\\$F := \forall v(f = v \rightarrow \exists {\bf x} (B_1(v,{\bf x}) \lor \dots \lor B_k(v,{\bf x})))$;
\\$G := \forall v(\exists {\bf x} (B_1(v,{\bf x}) \lor \dots \lor B_k(v,{\bf x})) \rightarrow f = v)$.

\item Eliminate variables from $F$ as follows:

\begin{enumerate}
\item $F := \exists {\bf x} (B_1(v,{\bf x})^v_f \lor \dots \lor B_k(v,{\bf x})^v_f)$ and then equivalently,
\\$F := \exists {\bf x} (B_1(v,{\bf x})^v_f) \lor \dots \lor \exists {\bf x}(B_k(v,{\bf x})^v_f)$.
\item $F_i := \exists {\bf x} (B_i(v,{\bf x})^v_f)$.
\item Eliminate variables from $F_i$ as follows:

\begin{enumerate}
\item $D_i := B_i(v,{\bf x})^v_f$.
\item While there is a variable $x$ still in $D_i$, select a conjunctive term $x = t$ or $t = x$ (such that no variable in $t$ depends on $x$) in $D_i$, then $D_i := (D_i)^x_t$.
\item $F_i = D_i$ (drop the existential quantifier since there are no variables in $D_i$).
\end{enumerate}

\item $F := F_1 \lor \dots \lor F_k$.
\end{enumerate}

\item Eliminate variables from $G$ as follows:

\begin{enumerate}
\item $G := \forall v{\bf x} ((B_1(v,{\bf x}) \lor \dots \lor B_k(v,{\bf x})) \rightarrow f = v)$ and then equivalently,
\\$G := \forall v{\bf x} (B_1(v,{\bf x})\rightarrow f = v) \land \dots \land \forall v{\bf x} (B_k(v,{\bf x}) \rightarrow f = v)$.
\item $G_i := \forall v{\bf x} (B_i(v,{\bf x})\rightarrow f = v)$.
\item Eliminate variables from $G_i$ as follows:

\begin{enumerate}
\item $D_i := B_i(v,{\bf x})\rightarrow f = v$.
\item While there is a variable $x$ still in $D_i$, select a conjunctive term $x = t$ or $t = x$ (such that no variable in $t$ depends on $x$) from the body of $D_i$, then $D_i := (D_i)^x_t$.
\item $G_i = D_i$ (drop the universal quantifier since there are no variables in $D_i$).
\end{enumerate}

\item $G := G_1 \lor \dots \lor G_k$.
\end{enumerate}

\item $E := F \land G$.
\end{enumerate}


The following proposition asserts the correctness of this method. Note that 
the absence of variables in arguments of uninterpreted functions can be achieved
by grounding ASP variables and enforcing that no SMT
variables occur in uninterpreted functions.

\begin{prop}\label{prop:elim}\optional{prop:elim}
For any completion of a variable isolated ASPMT
program with no variables in arguments of uninterpreted functions, 
applying variable elimination method repeatedly results in a  classically
equivalent formula that contains no variables. 

\end{prop}
\cbla

For example, given the equivalence (\ref{ex:comp}),
Step 2a) turns the implication from left to right into the formula
{\small
\[
\ba{rl}
\exists xd (& 
              ~~~~(\i{Accel}(0)\mvis\true\ \land \ 
               \i{Speed}(0)\mvis x\ \land\  
               \i{Duration}(0)\mvis d\ \land (\i{Speed}(1)\mvis x+{\rm A}\times d) ) \\
             & \lor \ 
              (\i{Decel}(0)\mvis\true\ \land \ 
               \i{Speed}(0)\mvis x\ \land\  
               \i{Duration}(0)\mvis d \land (\i{Speed}(1)\mvis x-{\rm A}\times d) ) \\
             & \lor \ 
               (\i{Speed}(0) = \i{Speed}(1)))
\ea
\]
}
And then step 2d) produces
{\small
\[
\ba{l}
(\i{Accel}(0)\mvis\true\land \i{Speed}(1)\mvis \i{Speed}(0) 
      + {\rm A}\times\i{Duration}(0)) \lor \\
(\i{Decel}(0)\mvis\true\land \i{Speed}(1) = \i{Speed}(0) - {\rm
  A}\times \i{Duration}(0)) \lor \\
(\i{Speed}(0) = \i{Speed}(1)).
\ea
\]
}
\BOCC
To see why variable isolation is required, consider the {\cblu formula} $f = X \leftrightarrow 2*X = Y \land 2*Y = X$. One step of 3c) produces the formula 
$$2*2*Y = Y \land X = X \rightarrow f = 2*Y$$. We can drop $X = X$ and then perform another step of 3c) to get $$2*2*(2*2*Y) = 2*2*Y \rightarrow f = 2*(2*2*Y)$$. Then, at this point, no conjunctive term exists of the 
form $x = t$ so the procedure terminates but the variable $Y$ still remains in the formula.
\EOCC

\BOCC
\begin{prop}\label{prop:elim2}\optional{prop:elim2}
Given the completion of an av-separated, variable isolated ASPMT theory, an equivalence from the completion $E = \forall v(f = v \leftrightarrow \exists {\bf x} (B_1({\bf x}) \lor \dots \lor B_k({\bf x})))$ where each $B_i$ is a conjunction of possibly negated atomic formulas and has $v = t$ as a non-negated subformula for some term $t$ not containing $v$, the result of applying the variable elimination method to $E$ contains no variables.
\end{prop}
\EOCC

\vspace{-0.5cm}
\section{{\sc aspmt2smt} System}\label{sec:system}

\subsection{Syntax of Input Language}

In addition to the syntactic restriction on ASPMT rules imposed in the previous section, the current version of system {\sc aspmt2smt} 
assumes that the input program is {\em ${\bf f}$-plain}
\cite{bartholomew12stable}, as well as ``av-separated,'' which intuitively 
means that no variable occurring in
an argument of an uninterpreted function is related to the value
variable of another uninterpreted functions via equality.\footnote{See
  the system homepage
  for the precise description of this condition.} For example, 
for the rule $f(x) = 1 \leftarrow g = y \land y = x$, variable $x$ is
an argument of $f$ and is also related to the value variable $y$ of $g$
via equality $y=x$.
The reason for this restriction is because the system sets the
equalities $g=y$ and $y = x$ aside (so that {\sc gringo} does not
ground them), and ground the rule and then
replace the equalities back to yield
\[
\ba l
f(1) = 1 \leftarrow g = y \land y = x \\
f(2) = 1 \leftarrow g = y \land y = x \\
\dots
\ea
\]
rather than the intended
\[
\ba l
f(1) = 1 \leftarrow g = y \land y = 1 \\
f(2) = 1 \leftarrow g = y \land y = 2 \\
\dots
\ea
\]
It should also be noted that the only background theories considered
in this version of the implementation are arithmetic over reals and
integers.



System {\sc aspmt2smt} uses a syntax similar to system {\sc cplus2asp}
\cite{babb13cplus2asp} for declarations and a syntax similar to
system {\sc f2lp}~\cite{leej09} for rules.

There are declarations of four kinds, {\tt sorts}, {\tt objects},
{\tt constants}, and {\tt variables}. The sort declarations specify user data
types (note: these cannot be used for value sorts). The object
declarations specify the elements of the user-declared data types. The
constant declarations specify all of the (possibly boolean) function
constants that appear in the theory. The variables declarations
specify the user-declared data types associated with each variable.
Declarations for the car example are shown below.

\lstset{
   basicstyle=\small\ttfamily,
   basewidth=0.55em,
   numbers=none,
   numberstyle=\tiny,
   stringstyle=\footnotesize\ttfamily,
   showspaces=false,
   showstringspaces=false
}

\begin{lstlisting}
:- sorts
  step; astep.

:- objects
  0..st           :: step;              0..st-1         :: astep.

:- constants
  time(step)      :: real[0..t];        accel(astep)    :: boolean;
  duration(astep) :: real[0..t];        decel(astep)    :: boolean;
  speed(step)     :: real[0..ms];       location(step)  :: real[0..l].
  
:- variables
  S               :: astep;             B               :: boolean.
\end{lstlisting}

Only propositional connectives are supported in this version of
{\sc aspmt2smt} and these are represented in the system as follows:
\smallskip
\begin{center}
\begin{tabular}{{|c|c|c|c|c|}} \hline
$~~\land$~~&~~$\lor$~~&~~$\neg$~~&~~$\rightarrow$~~&~~$\leftarrow$ \\ \hline
{\tt \&} & {\tt |} & {\tt not} & {\tt ->} & {\tt <-} \\ \hline
\end{tabular}
\end{center}

Comparison and arithmetic
operators are represented as usual: 

\begin{center}
\begin{tabular}{|c|c|c|c|c|c|c|c|c|c|} \hline
$~~<~~$ & $~~\leq~~$ & $~~\geq~~$ & $~~>~~$ & $~~=~~$ & $~~\neq~~$ & ~add~ & subtract & multiply & divide \\
\hline
{\tt <} & {\tt <=} & {\tt >=} & {\tt >} & {\tt =} & {\tt !\!=} & {\tt
  +} & {\tt -} & {\tt *} & {\tt /} \\ \hline
\end{tabular}
\end{center}

$a\  !\!= b$ is understood as $\neg (a = b)$. To abbreviate the
formula $A \lor \neg A$, which is useful for expressing defaults and
inertia, we write $\{A\}$. The rest of the car example is shown below.

\begin{lstlisting}
% Actions and durations are exogenous
{accel(S)=B}.
{decel(S)=B}.
{duration(S)=X}.

% no concurrent actions 
<- accel(S)=true & decel(S)=true.

% effects of accel and decel
speed(S+1)=Y <- accel(S)=true & speed(S)=X & duration(S)=D & Y = X+ar*D. 
speed(S+1)=Y <- decel(S)=true & speed(S)=X & duration(S)=D & Y = X-ar*D. 

% preconditions of accel and decel
<- accel(S)=true & speed(S)=X & duration(S)=D & Y = X+ar*D & Y > ms. 
<- decel(S)=true & speed(S)=X & duration(S)=D & Y = X-ar*D & Y < 0. 

% inertia of speed
{speed(S+1)=X} <- speed(S)=X.

location(S+1)=Y <- location(S)=X & speed(S)=A & 
    speed(S+1)=C & duration(S)=D & Y = X+(A+C)/2*D.

time(S+1)=Y <- time(S)=X & duration(S)=D & Y=X+D.

% problem instance
time(0)=0.     speed(0)=0.     location(0)=0.
<- location(st) = Z & Z != l.
<- speed(st) = Z & Z != 0.
<- time(st) = Z & Z != t.
\end{lstlisting}

This description can be run by the command 
\begin{lstlisting}
    $aspmt2smt car -c st=3 -c t=4 -c ms=4 -c ar=3 -c l=10
\end{lstlisting}
which yields the output
\begin{lstlisting}
  accel(0) = true  accel(1) = false  accel(2) = false
  decel(0) = false  decel(1) = false  decel(2) = true
  duration(0) = 1.1835034190  duration(1) = 1.6329931618
  duration(2) = 1.1835034190  location(0) = 0.0
  location(1) = 2.1010205144  location(2) = 7.8989794855
  location(3) = 10.0  speed(0) = 0.0
  speed(1) = 3.5505102572  speed(2) = 3.5505102572
  speed(3) = 0.0  time(0) = 0.0  time(1) = 1.1835034190
  time(2) = 2.8164965809  time(3) = 4.0
z3 time in milliseconds: 30
Total time in milliseconds: 71
\end{lstlisting}

\subsection{Architecture}

\begin{figure}
\centering
 \includegraphics[width=0.9\textwidth]{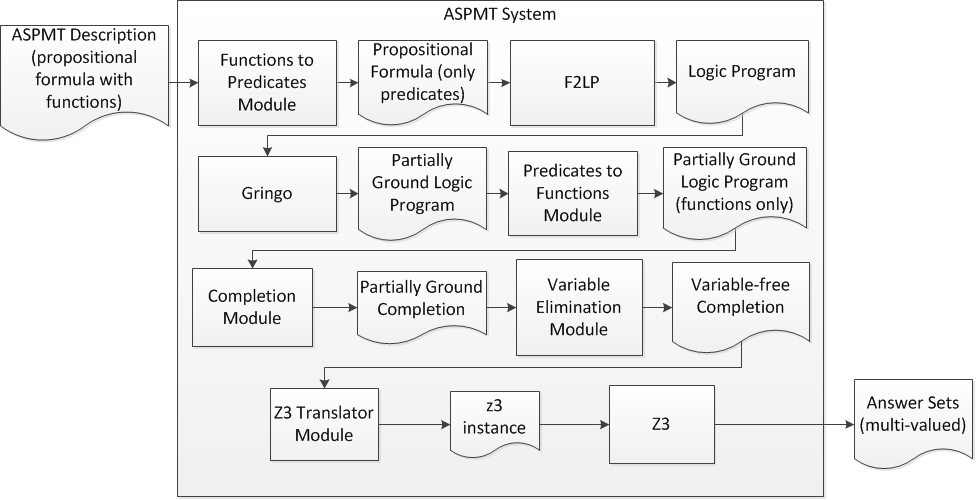}
\caption{{\sc aspmt2smt} System Architecture}
\label{fig:architecture}
\end{figure}
%
The architecture of {\sc aspmt2smt} system is shown in Figure~\ref{fig:architecture}.
The system first converts the ASPMT description to a
propositional formula containing only predicates. In addition, this
step substitutes auxiliary constants for SMT variables and necessary
preprocessing for {\sc f2lp} (v1.3) and {\sc gringo} (v3.0.4) to enable partial grounding of ASP variables only. {\sc f2lp} transforms the propositional formula
into a logic program and then {\sc gringo} performs partial grounding
on only the ASP variables. The {\sc aspmt2smt} system then
converts the predicates back to functions and replaces the 
auxiliary constants with the original expressions. Then the system
computes the completion of this partially ground logic program and
performs variable elimination on that  completion. Finally, the system
 converts this variable-free description into the language of {\sc
  z3} and then relies on {\sc z3} to produce models which correspond
to stable models of the original ASPMT description.

For instance, consider the result of variable elimination on the 
portion of the completion related to $speed(1)$ of the running car example:
\[
\small
\ba{l}
(\i{Accel}(0)\mvis\true\land \i{Speed}(1)\mvis \i{Speed}(0) 
      + {\rm A}\times\i{Duration}(0)) \lor \\
(\i{Decel}(0)\mvis\true\land \i{Speed}(1) = \i{Speed}(0) - {\rm
  A}\times \i{Duration}(0)) \lor \\
(\i{Speed}(0) = \i{Speed}(1)).
\ea
\]
In the language of {\sc z3}, this is
\begin{lstlisting}
(assert (or (or 
   (and (= accel_0_ true) (= speed_1_ (+ speed_0_ (* duration_0_ a))))
   (and (= decel_0_ true) (= speed_1_ (- speed_0_ (* duration_0_ a)))))
   (= speed_1_ speed_0_) ))
\end{lstlisting}

\section{Experiments}\label{sec:experiment}

The following experiments demonstrate the capability of the {\sc aspmt2smt}
system to perform nonmonotonic reasoning about continuous changes. In addition, this shows a significant performance increase compared to ASP solvers for domains in which only SMT variables have large domains. However, when ASP variables have large domains, similar scalability issues arise as comparable grounding still occurs. 


All experiments were performed on an Intel Core
2 Duo 3.00 GHZ CPU with 4 GB RAM running Ubuntu 13.10.
The domain descriptions of these examples can be found from the system
homepage. 

\subsection{Leaking Bucket}

\begin{wrapfigure}{l}{0.6\textwidth}
{\footnotesize
\ \\\begin{tabular}{|c|c|c|}
\hline 
c & {\sc clingo} v3.0.5 & {\sc aspmt2smt} v0.9 \\ 
 &  Run Time   & Run Time  \\  
 & (Grounding + Solving) & (Preprocessing + solving)   \\\hline\hline
10 & 0s (0s+0s) & .037s (.027s + .01s) \\\hline
50 & .02s (02s + 0s) & .089s (.079s + .01s) \\\hline
100 & .12s (.12s + 0s) & .180s (.170s + .01s) \\\hline
500 & 8.69s (8.68s + .01s) & 1.731s (1.661s + .07s) \\\hline
1000 & 60.32s (60.29s+ .03s) & 35.326s (35.206s + .12s) \\\hline
\end{tabular}
}
\end{wrapfigure}

Consider a leaking bucket with maximum capacity $c$ that loses one
unit of water every time step by default. The bucket can be refilled
to its maximum capacity by the action $fill$. The initial capacity is
5 and the desired capacity is 10.
\BOCC
\begin{lstlisting}
:-sorts
  atime;time.

:-objects
  0..c :: step;
  0..c-1  :: astep.

:-constants
  amt(step) :: int[0..c];
  fill(astep) :: boolean.

:-variables
  T :: step;
  ST :: astep;
  X :: int[0..c].

{amt(ST+1) = X-1} <- amt(ST) = X.
{fill(ST) = true}.
{fill(ST) = false}.

amt(ST+1) = X <- fill(ST) = true & X = c.
<- amt(T) = X & X < 2.

amt(0) = 5.
<- not (amt(c) = 10).
\end{lstlisting}
\EOCC

We see that in this experiment, {\sc aspmt2smt} does not yield
significantly better results than {\sc clingo}. The reason for this
is that the scaling of this domain takes place in the number of time steps. Thus, since {\sc aspmt2smt} uses {\sc gringo} to generate fluents 
for each of these time steps, the ground descriptions given to {\sc clingo} and {\sc z3} are of similar size. Consequently, we see that the majority
of the time taken for {\sc aspmt2smt} is in preprocessing.

\subsection{Car Example}

\begin{wrapfigure}{l}{0.6\textwidth}
{\footnotesize
\begin{tabular}{|c|c|c|}
\hline 
k & {\sc clingo} v3.0.5  & {\sc aspmt2smt} v0.9 \\ 
 &  Run Time   & Run Time  \\ 
 & (Grounding + Solving) & (Preprocessing + solving) \\ \hline\hline
1 & n/a & .084s (.054s + .03s) \\\hline
5 & n/a & .085s (.055s + .03s) \\\hline
10 & n/a & .085s (.055s + .03s) \\\hline
50 & n/a & .087s (.047s + .04s) \\\hline
100 & n/a & .088s (.048s + .04s) \\\hline\hline

1 & .61s (.6s + .01s) & .060s (.050s + .01s) \\\hline
2 & 48.81s (48.73s + .08s) & .07s (.050s + .02s) \\\hline
3 & $>30$ minutes & .072s (.052s + .02s) \\\hline
5 & $>30$ minutes & .068s (.048s + .02s)\\\hline
10 & $>30$ minutes & .068s (.048s + .02s) \\\hline
50 & $>30$ minutes & .068s (.048s + .02s)  \\\hline
100 & $>30$ minutes & .072s (.052s + .02s) \\\hline
\end{tabular}
}
\end{wrapfigure}

Recall the car example in Section~\ref{ssec:aspmt}. 
\BOCC
A car is on a road of length L.  If the accelerator is activated, the
car will speed up with constant acceleration A until the accelerator is
released or the car reaches its maximum speed MS, whichever comes first.
If the brake is activated, the car will slow down with acceleration -A
until the brake is released or the car stops, whichever comes first.
Otherwise, the speed of the car remains constant.
Give a formal representation of this domain, and write a program that
uses your representation to generate a plan satisfying the following
conditions:  at duration 0, the car is at rest at one end of the road; at
duration T, it should be at rest at the other end.
\EOCC
The first half of the experiments are done with the values L = 10$k$, A = 3$k$, MS = 4$k$, T = 4$k$, which yields solutions with irrational values and so cannot be solved by system {\sc clingo}.
The second half of the  experiments are done with the values L = 4$k$, A = $k$, MS = 4$k$, T = 4$k$, which yields solutions with integral values and so can be solved by system {\sc clingo}. In this example, only the SMT variables have increasing domains but the ASP variable domain remains the same. Consequently, the {\sc aspmt2smt} system scales very well compared to the ASP system which can only complete the two smallest size domains.

We also experimented with {\sc clingcon}. Since {\sc clingcon} does
not allow intensional functions, we need to encode the example
differently using auxiliary abnormality atoms to represent the notions
of inertia and default behaviors. In the first set of experiments,
{\sc clingcon} performed better than {\sc aspmt2smt}, but
like {\sc clingo}, the current version of {\sc clingcon} cannot handle real numbers, so it is not applicable to the second set of experiments.


\subsection{Space Shuttle Example}

\BOCC
\begin{lstlisting}
:- sorts
   step;astep;axis.

:- objects
   0..st   :: step;
   0..st-1 :: astep;
   x,y,z   :: axis.

:- constants
   duration(astep)      :: real[0..2];
   time(step)           :: real[0..2];
   mass                 :: real[0..m];
   speed(axis,step)    :: real[0..v];
   pos(axis,step)      :: real[0..p];
   jet1fire(axis,astep) :: real[0..f];
   jet2fire(axis,astep) :: real[0..f].

:- variables
  S      :: step;
  AS     :: astep;
  AX,AX1 :: axis.

mass = m.

speed(x,0) = 0.
speed(y,0) = 1.
speed(z,0) = 1.

time(0) = 0.

pos(x,0) = 0.
pos(y,0) = 0.
pos(z,0) = 0.

{duration(AS) = X}.
{jet1fire(AX,AS) = X}.
{jet2fire(AX,AS) = X}.

<- jet1fire(AX,AS) = X & jet1fire(AX1,AS) = X1 & 
    X != 0 & X1 != 0 & AX != AX1.
<- jet2fire(AX,AS) = X & jet2fire(AX1,AS) = X1 & 
    X != 0 & X1 != 0 & AX != AX1.

pos(AX,AS+1) = Z <- pos(AX,AS) = X & duration(AS) = T & 
   speed(AX,AS) = S0 & speed(AX,AS+1) = S1 & 
	 Z = X + T*(S0+S1)/2.
speed(AX,AS+1) = Z <- jet1fire(AX,AS) = X1 & 
   jet2fire(AX,AS) = X2 & duration(AS) = T & mass = M & 
	 speed(AX,AS) = Y & Z = Y+T*(X1+X2)/M.

time(AS+1) = X <- time(AS) = Y & duration(AS) = T & 
    X = Y+T.

<- pos(x,st) = X & X != 0.
<- pos(y,st) = X & X != 3*k.
<- pos(z,st) = X & X != 2*k.

\end{lstlisting}
\EOCC

\begin{wrapfigure}{l}{0.6\textwidth}
{\footnotesize
\begin{tabular}{|c|c|c|}
\hline 
k & {\sc clingo} v3.0.5 & {\sc aspmt2smt} v0.9  \\ 
 &  Run Time   & Run Time  \\ 
 & (Grounding + Solving) & (Preprocessing + solving) \\ \hline\hline
1 & 0s (0s + 0s)& .048s (.038s + .01s) \\\hline
5 & .03s (.02s + .01s) & .047s (.037s + .01s) \\\hline
10 & .14s (.9s + .5s) & .053s (.043s + .01s) \\\hline
50 & 7.83s (3.36s + 4.47s) & .050s (.040s + .01s) \\\hline
100 & 39.65s (16.14s + 23.51s) & .051s (.041s + .01s) \\\hline
\end{tabular}
}
\end{wrapfigure}

The following example is from~\cite{lee03}, which represents
cumulative effects on continuous changes. A spacecraft is not
affected by any external forces. It has two jets and the force that
can be applied by each jet along each axis is at most 4$k$. The initial
position of the rocket is (0,0,0) and its initial velocity is
(0,1,1). How can it get to (0,3$k$,2$k$) within 2 seconds? Assume the mass
is 2.

Again in this problem, the scaling lies only in the size of the value of the functions 
involved in the description. Consequently, we see
no scaling issues in either {\sc aspmt2smt} or {\sc clingcon}.

\subsection{Bouncing Ball Example}

\begin{wrapfigure}{l}{0.6\textwidth}
\ \\{\footnotesize
\begin{tabular}{|c|c|c|}
\hline 
k & {\sc clingo} v3.0.5 & {\sc aspmt2smt} v0.9 \\ 
 &  Run Time   & Run Time  \\ 
 & (Grounding + Solving) & (Preprocessing + solving) \\ \hline\hline
1 & n/a & .072s (.062s + .01s) \\\hline
10 & n/a &.072s (.062s + .01s) \\\hline
100 & n/a &.071s (.061s + .01s) \\\hline
1000 & n/a &.075s (.065s + .01s) \\\hline
10000 & n/a &.082s (.062s + .02s) \\\hline
\end{tabular}
}
\end{wrapfigure}

The following example is from~\cite{chintabathina08towards}. 
A ball is held above the ground by an agent. The actions available to
the agent are {\sl drop} and {\sl catch}. Dropping the ball causes the height of
the ball to change continuously with time as defined by Newton's laws
of motion. As the ball accelerates towards the ground it gains
velocity. If the ball is not caught before it reaches the ground, it
hits the ground with speed $s$ and bounces up into the air with speed
$r \times s$ where $r = .95$ is the rebound coefficient. The bouncing ball
reaches a certain height and falls back towards the ground due to
gravity. An agent is holding a ball at height 100$k$. We want to have
the ball hit the ground and caught at height 50.

\BOCC
\begin{lstlisting}
:- sorts
   step;astep.

:- objects
   0..st   :: step;
   0..st-1 :: astep.

:- constants
   pos(step)   :: real[0..p];
   speed(step) :: real[-5000..5000];
   drop(astep)  :: boolean;
   catch(astep) :: boolean;
   duration(astep) :: real[0..1000];
   gravity     :: real[-50..50];
   coefficient :: real[0..1];
   holding(step) :: boolean.

:- variables
  S  :: step;
  AS :: astep.

coefficient = 95/100.
gravity = -98/10.

pos(0) = p.
holding(0) = true.
speed(0)  = 0.

{duration(AS) = X}.
{drop(AS) = true}.
{drop(AS) = false}.
{catch(AS) = true}.
{catch(AS) = false}.

<- drop(AS) = true & catch(AS) = true.
<- drop(AS) = true & holding(AS) = false.
<- catch(AS) = true & holding(AS) = true.

<- drop(AS) = true & duration(AS) = X & X != 0.
<- catch(AS) = true & duration(AS) = X & X != 0.

holding(AS+1) = true <- catch(AS) = true.
speed(AS+1) = 0 <- catch(AS) = true.
holding(AS+1) = false <- drop(AS) = true.

holding(AS+1) = true <- holding(AS) = true & 
    drop(AS) = false.
holding(AS+1) = false <- holding(AS) = false & 
    catch(AS) = false.

{speed(AS+1) = X} <- speed(AS) = Y & duration(AS) = T & 
    gravity = G & X = Y+T*G & holding(AS) = false.
speed(AS+1) = X <- speed(AS) = X & holding(AS) = true.
speed(AS+1) = X <- speed(AS) = Y & coefficient = C & 
    X = -1*Y*C & pos(AS) = 0 & holding(AS) = false.

<- pos(S) = X & X < 0.

pos(AS+1) = X <- pos(AS) = Y & duration(AS) = T & 
    speed(AS+1) = S2 & speed(AS) = S1 &	
		X = Y + T*(S1+S2)/2 & ((catch(AS) = false & 
		holding(AS) = false) | drop(AS) = true).
pos(AS+1) = X <- pos(AS) = X & 
    ((holding(AS) = true & drop(AS) = false) | 
		catch(AS) = true).

<- pos(st-2) = X & X != 0.
<- pos(st) = X & X != 50.
\end{lstlisting}
\EOCC

Again, {\sc clingo} and {\sc clingcon} are unable to find solutions to this 
domain since solutions are not integral. Also, we see that {\sc aspmt2smt} suffers
no scaling issues here again due to the fact that in this problem the scaling lies 
only in the size of the value of the functions involved in the description.


\section{Conclusion}

We presented system {\sc aspmt2smt}, which translates ASPMT instances
into SMT instances, and uses SMT solvers to compute ASPMT. 
Unlike other ASP solvers, this system can compute effective real
number computation by leveraging the effective SMT solvers. 
Future work includes extending the system to handle other background
theories, and investigate a larger fragment of ASPMT instances that
can be turned into SMT instances. 

\smallskip\noindent
{\bf Acknowledgements}\ \ 
We are grateful to the anonymous referees for their useful comments. 
This work was partially supported by the National Science Foundation under Grant IIS-1319794 and by the South Korea IT R\&D program MKE/KIAT
2010-TD-300404-001.

\bibliographystyle{splncs}



\end{document}